\documentclass[]{llncs}
\usepackage{llncsdoc}

\usepackage{graphics} 
\usepackage{graphicx}
\usepackage[caption=false,font=footnotesize]{subfig}
\usepackage{color}
\usepackage{epsfig} 
\usepackage[noadjust]{cite}
\usepackage{import}

\usepackage[breaklinks,colorlinks=true,linkcolor=black,citecolor=black,urlcolor=black]{hyperref}

\graphicspath{{./assets/}}

\title{
	RAFCON: a Graphical Tool for\\Task Programming and Mission Control
}
\titlerunning{RAFCON: a Graphical Tool for Task Programming and Mission Control}

\author{
	Sebastian G. Brunner\thanks{Both authors contributed equally to this work.}\inst{1} \and
	Franz Steinmetz$^\star$\inst{1} \and
	Rico Belder\inst{1} \and
	Andreas D\"omel\inst{1}
}
\authorrunning{Brunner and Steinmetz et al.}

\institute{
	$^1$Robotics and Mechatronics Center (RMC) of the German Aerospace Center (DLR), Oberpfaffenhofen-Wessling, Germany\\
	\email{firstname.lastname@dlr.de}
}

\begin{document}

\maketitle

\begin{abstract}

	There are many application fields for robotic systems including service robotics, search and rescue missions, industry and space robotics. As the scenarios in these areas grow more and more complex, there is a high demand for powerful tools to efficiently program heterogeneous robotic systems. Therefore, we created RAFCON, a graphical tool to develop robotic tasks and to be used for mission control by remotely monitoring the execution of the tasks. To define the tasks, we use state machines which support hierarchies and concurrency. Together with a library concept, even complex scenarios can be handled gracefully. RAFCON supports sophisticated debugging functionality and tightly integrates error handling and recovery mechanisms. A GUI with a powerful state machine editor makes intuitive, visual programming and fast prototyping possible. We demonstrated the capabilities of our tool in the SpaceBotCamp national robotic competition, in which our mobile robot solved all exploration and assembly challenges fully autonomously. It is therefore also a promising tool for various RoboCup leagues.

\end{abstract}


\section{Introduction}
\label{sec:introduction}
Managing the heterogeneous modules (e.\,g. navigation, vision, manipulation etc.) of a robot is challenging, as the scenarios in common robotic application fields like household and industries grow more and more complex. In this work, we thus focus on how complex tasks can be programmed and how all subsystems of a robot orchestrated at a central instance using visual programming and hierarchical state machines.

For solving complex tasks, one approach is to semantically specify the robot and its environment in a planning domain on which a task planner can be used to infer all steps for reaching a certain goal, which is also specified in the planning domain. PDDL~\cite{pddl:1998} is a common solution for such planning problems and is often used in service robotic scenarios \cite{leidner2012things}.
Such planners often suffer from over- or under-constraint models and fail if a real world failure cannot be represented in their environment model \cite{bohren2010smach}. Furthermore, they have a much higher computational footprint. Therefore, they are not suited for many real-world tasks, e.\,g. industrial scenarios.

Alternatively, state machines \cite{krithivasan2014theory} often come into play to specify the behavior of the robot in a more bottom-up like approach (see \cite{bohren2011towards, jentzsch2013tums, nguyen2013ros}). The robotic system is always in a certain state and proceeds to the next state depending on internal or external events. As classical state machines have problems coping with complex scenarios, powerful dialects were invented like \textit{statecharts}~\cite{harel1987statecharts} and \textit{SyncCharts}~\cite{andre1995synccharts}. They augment a classical state machine with hierarchy and concurrency concepts, preemption handling, error recovery and data management. Our state machine dialect uses and adapts many of these features and is furthermore based on \textit{flowcharts} in regards of its eventless design.

Specifically for programming robotic tasks, there exist many tools. Next to well-designed solutions for educational purposes like \textit{Scratch}~\cite{resnick2009scratch} or \mbox{\textit{NXT-G}}~\cite{kelly2010lego} from LEGO Mindstorms, all tools designed for real world robots suffer from certain problems. For some tools, maintenance and support was canceled, e.\,g. \textit{ROS Commander} \cite{nguyen2013ros}, \textit{RobotFlow} \cite{cote2004code} or \textit{MissionLab}~\cite{arkin2006missionlab}, others do not offer their code to the open source community, like \textit{Gostai Studio}~\cite{baillie2008urbi}, or do not provide a graphical editor, such as \textit{SMACH}~\cite{bohren2010smach}.

Therefore, we developed RAFCON, a visual programming tool, allowing for the creation of hierarchical state machines. It is written in Python, as the language is interpreted, easy to learn and can be integrated with software modules of other languages. 
RAFCON was created completely from scratch. It is inspired by the flow control tool \textit{Bubbles}, which has been developed at our institute some years ago~\cite{widmoser2012interaction}. Before starting with the implementation, experienced roboticist of our institute elaborated a long list of requirements. The key advantages of our tool are the novel visualization supporting state machines of several hundreds of states (see Sec.~\ref{sec:experiments}), powerful error recovery mechanisms, sophisticated debugging functionalities and usability and intuitiveness to allow for fast prototyping.

\begin{figure}[!htb]
\minipage{0.5\textwidth}
  \centering
  \includegraphics[scale=0.15]{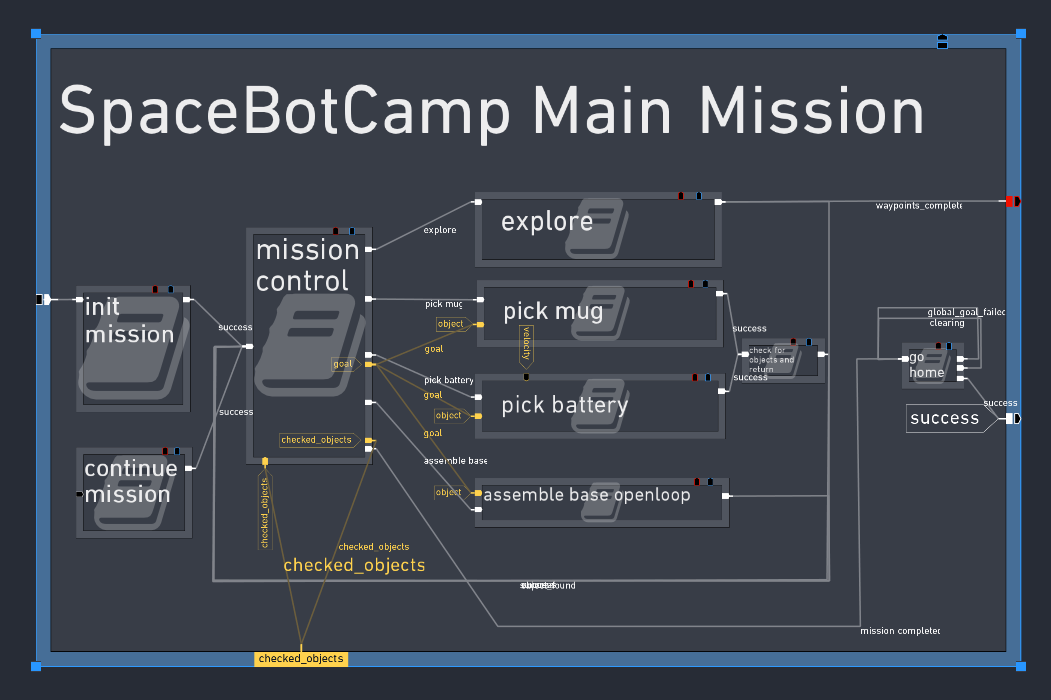}
  \label{fig:sbc_rafcon}
\endminipage
\minipage{0.5\textwidth}
  \centering
  \includegraphics[scale=0.15]{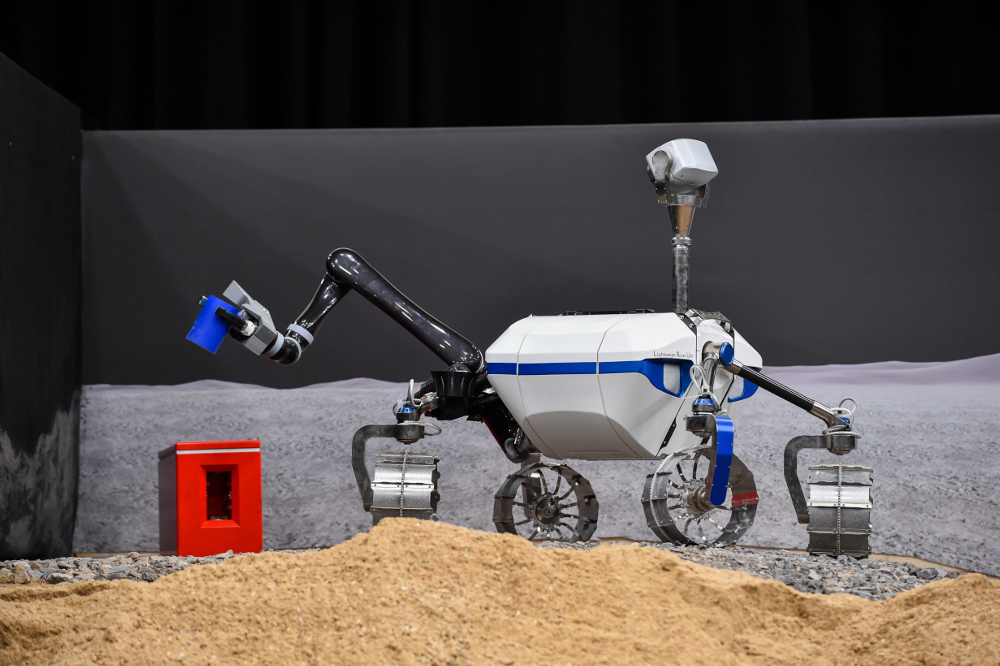}
  \label{fig:lru_sbc2}
\endminipage
\caption{The left figure shows the state machine for the SpaceBotCamp, in which we took part with our Lightweight Rover Unit (LRU, \cite{wedler2015lru}). On the right, the LRU is depicted while it mounts a blue container onto the red base station.
}
\label{fig:rafcon_lru_sbc}
\vspace{-1.5em}
\end{figure}

Furthermore, RAFCON enables collaborative state machine development. During the SpacebotCamp 2016\footnote{http://s.dlr.de/ura7}, we successfully used RAFCON as both an autonomous task control software on a mobile robot as well as part of a mission control center setup with powerful remote monitoring and control capabilities. The mission in the SpacebotCamp included autonomous exploration and localization on a moon-like terrain, as well as object detection and assembly, all within a 60\,minute time limit (see Fig.~\ref{fig:rafcon_lru_sbc}). Thereby, many challenges had to be tackled that are also common e.\,g. in the RoboCup Rescue League.

The paper is structured as follows: At first we explain the core framework in Sec.~\ref{sec:core_framework}. The following Sec.~\ref{sec:gui} describes all important components of the GUI. After proofing the capabilities of our task programming tool in the case study in Sec.~\ref{sec:experiments}, we will summarize our results and future work in Sec.~\ref{sec:conclusion}.


\section{Core framework}
\label{sec:core_framework}	
The core of the RAFCON framework mainly consists of the \textit{state machine} and the \textit{execution engine}. Hereby, next to the logical flow, also data flow concepts are supported. All of these concepts are described in the following.

\paragraph{State machine:} A state machine contains an execution engine and a \textit{root state} that is the starting point of the execution. State machines are hierarchical, meaning that there are states that can contain child states.

\paragraph{State:} The states of a state machine are the instances at which actions take place (Moore machine). There are four different types of states:

\begin{itemize}
	\item \textit{Execution states} are the essential states as they contain a user-defined function called \verb|execute|, which is written in Python code. This \verb|execute| function serves as connection to other middleware. These states cannot have any children.
	\item \textit{Hierarchy states} group several child states. Each hierarchy state has a fixed \textit{start state} that defines the entry point of the state.
	\item \textit{Concurrency states} also group several child states, however all of these child states are evaluated concurrently, i.\,e. in parallel. The subtype \textit{preemptive concurrency state} stops all child states, as soon as one child state has finished its execution, while a \textit{barrier concurrency state} waits for all child states to finish.
	\item \textit{Library states} are intended to reuse state machines. They simply wrap a whole state machine.
\end{itemize}

\paragraph{Outcome:} Each state has two or more outcomes. As the name implies, these elements define the possible exit statuses of a state. Mandatory outcomes are \textit{aborted}, in case an error occurred in the state, and \textit{preempted}, for which the state was preempted from outside.

\paragraph{Transition:} States are connected via transitions. A transition starts at an outcome and either ends in a sibling state or in an outcome of a parent state. If a state ends with a certain outcome, the transition connected to that outcome is followed to determine the next state.

\paragraph{Data port:} Next to outcomes, states can also have \textit{input data ports} and \textit{output data ports}. They correspond to parameters respectively return values of functions. Data ports have a name, type and default value.

\paragraph{Data flow:} Data ports of the same type can be connected using data flows. Thereby, the value assigned to the source ports gets forwarded to the target port.

\paragraph{Execution engine:} The execution engine runs a state machine, starting at the root state. If a hierarchy state is reached, the execution goes down in the hierarchy to the defined start state. The execution can split up, if a concurrency state is executed. The hierarchy is went back up, if a transition is reached that goes from a child to its parent state. For execution states, the values assigned to the input data ports are forwarded to the \verb|execute| function that can contain arbitrary code. The function can also assign values to the output data ports and defines the outcome of the state. 

Many features of the engine help in debugging and testing even complex state machines. The engine supports continuous and step mode. In step mode, one can step over, into and out of states. In addition, step back mode is possible, for which execution states can have a separate \verb|execute_backwards| function. One can also command the engine to start execution at an arbitrary state. During the execution, a state machine can be changed on the fly. Finally, the execution can be controlled from a separate host, which is especially useful for mobile robots.



\section{Graphical user interface}
\label{sec:gui}	
\begin{figure*}[!htb] 
	\centering
	\includegraphics[width=\textwidth]{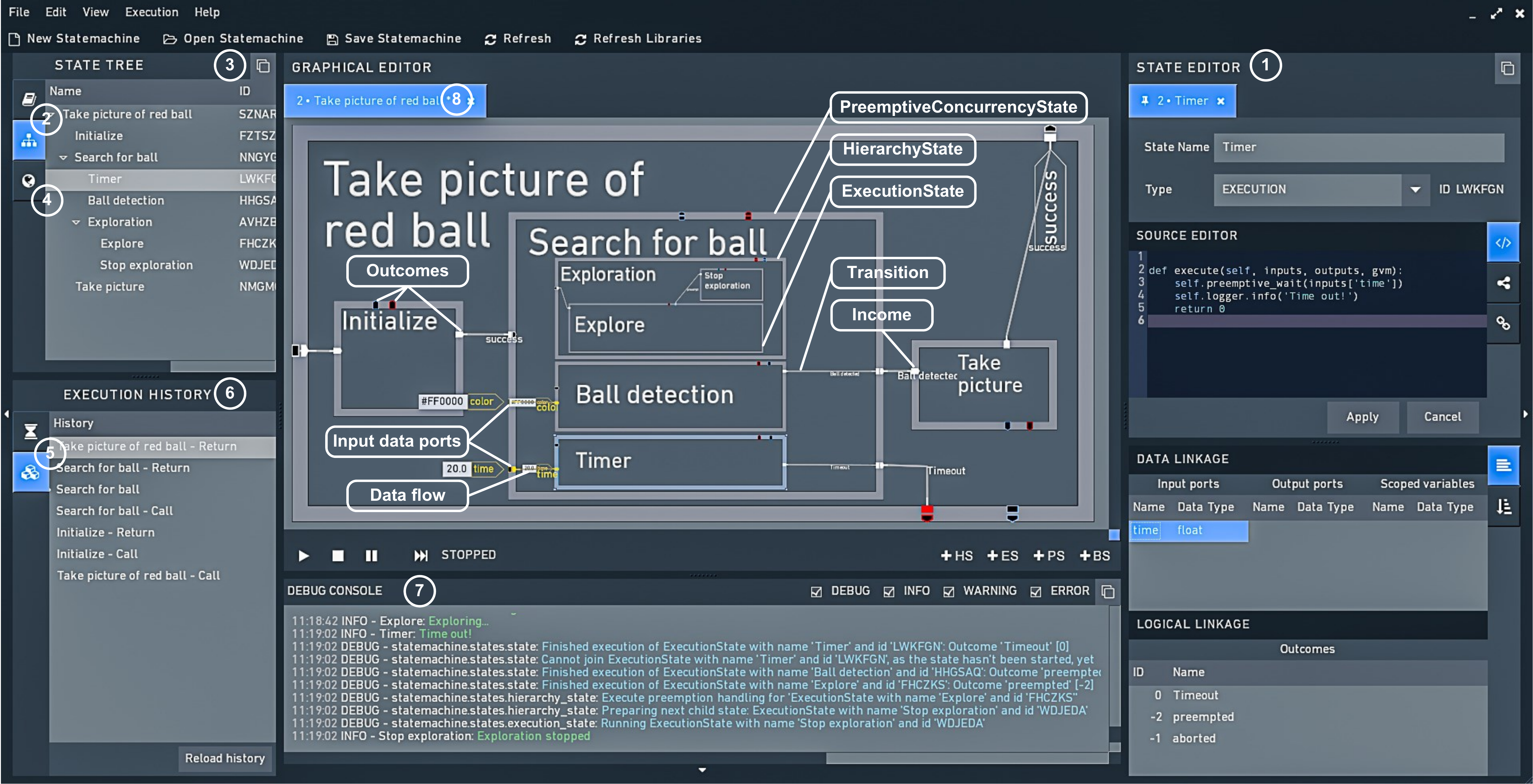}
	\caption{An example state machine visually programmed in RAFCON. An autonomous agent explores a unknown environment until it localizes a red ball or stops because a timer preempts the execution.}
	\label{fig:rafcon_gui}
	\vspace{-1.5em}
\end{figure*}

The graphical user interface, shown in Fig.~\ref{fig:rafcon_gui}, is the most prominent feature of RAFCON. The central widget, the  \textit{Graphical State Machine Editor}, is the part of the GUI that renders RAFCON unique amongst other (visual) task programming tools. Next to the creation of state machines, this GUI enables the user to execute and monitor the state machine, also from remote. The GUI design was developed in cooperation with professional interface designers\footnote{Interaktionswerk, https://interaktionswerk.de/}.

Concerning task engineering, visual programming is in our opinion superior to textual languages in this context due to several reasons: First of all, the overview and the understanding of a state machine is increased as the user builds a mental model~\cite{navarro-prieto2001are} of it. Beyond that, the logic and data flows are visualized separately, which gives a clear view on the data handling and routing. Furthermore, visual programming is more intuitive. All this improves the speed of state machine creation.

\subsection{GUI layout}

The GTK+ widget toolkit was used to implement the GUI. A model-view-controller architecture is used for the GUI in order to communicate with the core. Thus, the core and the GUI are clearly separated. The four \textit{Gestalt principles}~\cite{chang2002gestalt} closure, similarity, continuity and proximity are heavily used to enhance information retrieval. A modular and flexible layout is achieved by making all sidebar tabs detachable, scalable and foldable.



All the different components of the GUI can be seen in Fig.~\ref{fig:rafcon_gui}. The right sidebar shows the \textit{state editor} (1). It lists all details of a state, like name, descriptions, ports and connections and offers functionality to modify all of these properties. Execution states have in addition a source view to edit their \verb|execute| function.

The left sidebar features many widgets with different purposes. A \textit{Library manager} (2) organizes all library states in a clear fashion for easy reuse. The \textit{State machine tree} (3) shows the structure of the state machine in a tree and can be used to explicitly select and navigate to a certain state. (4) is the \textit{Global variable manager} managing all global variables, (5) shows the \textit{Modification history} of all changes performed to the a state machine under construction and the \textit{Execution history} (6) keeps track of all states during an execution including the context data.

The sidebar (7) at the bottom is the \textit{Logging View}. Here all output of the executed states, the core and the GUI is collected and can be filtered by their logging level.

\subsection{Graphical editor}
\label{sec:Graphical_editor}

As already mentioned, the Graphical State Machine Editor is the most sophisticated element of the GUI. A clear visualization of complex state machines with highly nested hierarchies (see Sec.~\ref{sec:experiments}) is a big challenge. We implemented a mature navigation solution that is often experienced in digital maps. Zooming into the state machine (e.\,g. with the mouse wheel) reveals more of the details of lower hierarchy levels, while zooming out hides their details. The panning mechanism enables the user to translate the view to another position.
Thus, the different hierarchy levels can be shown in varying degrees of detail, depending on the states the user is interested in.

The editor can also be used for direct interaction with the state machine for e.\,g. handling states, creating connections with via-points, copying and pasting of states and moving logic and data ports along the state border.
Different view modes help the user to focus on the kind of information he is interested.



\section{Case study}
\label{sec:experiments}

In November 2015, the DLR Space Administration organized the SpaceBotCamp 2016 with the aim to stimulate research and innovations in autonomous space robotic scenarios. The mission of the competition was to explore a rough, unknown terrain with a mobile robot, to localize and pick up two objects and finally to assemble those objects on a base station located on a crushed stone hill top, shown in Fig.~\ref{fig:rafcon_lru_sbc}. Next to having no uplink to the robot for the majority of the time, and complete communication blackouts for two occasions, there was also a constant two second communication delay. Concerning these restrictions, a highly autonomous behavior was required from the LRU robot~\cite{schuster2016thelru} to solve the challenge in less than 60 minutes.

\begin{figure}
	\centering
	\includegraphics[width=0.35\columnwidth]{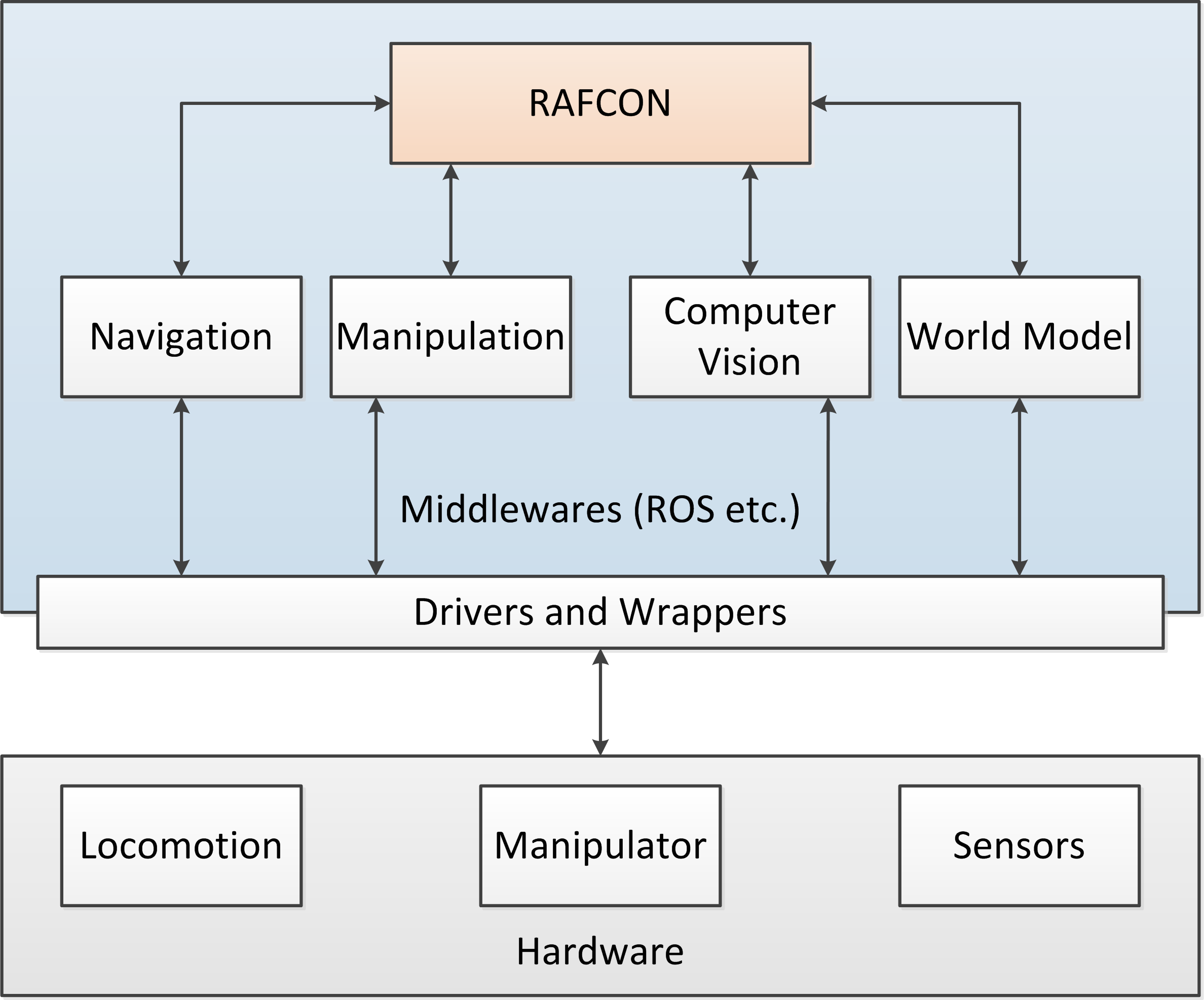}
	\caption{The architecture in the SpaceBotCamp, in which RAFCON is coordinating the main software modules of our mobile robot.}
	\label{fig:sbc_overview}
	\vspace{-1.5em}
\end{figure}

Our robot accomplished all tasks in only half of the given time limit. Hereby, RAFCON played a central role in autonomously orchestrating all modules of the system, like navigation, manipulation and vision. An abstract overview of this system architecture is shown in Fig.~\ref{fig:sbc_overview}.

RAFCON enabled us to collaboratively define the first two hierarchies of the overall state machine in the beginning. Subsequently, the state machine could be clearly divided into several sub-state-machines that were programmed by different developers. All in all, our final state machine consisted of more than 750 states and more than 1200 transitions. The maximum depth, i.\,e. number of hierarchies, was eight. This is a high number, considering the rather high-level nature of the used states.

During competition runtime, a ground station team was allowed to monitor the robot. Hereby, we were able to observe the current status of the state machine execution remotely. Therefore, several ground stations operators could subscribe to the state machine running on the robot and were supplied with status data of the state machine. This included the current execution point(s) and the values of data ports. Thus, an operator for navigation could observe the correct behavior of the robot during navigation, a manipulation expert could examine correct object assembly and an operator for computer vision could keep track of the detection and pose estimation procedures.



\section{Conclusions}
\label{sec:conclusion}
The visual programming tool RAFCON, presented in this paper, fills a gap in the robot programming domain. While for example ROS~\cite{quigley2009ros} unifies the communication in a heterogenous system, there is currently no graphical tool at hand for using that communication. RAFCON seamlessly integrates with ROS or other middlewares to orchestrate the different components in a way that they together perform a certain task. Our tool features a clear programming interface (API) that can be used for programmatic state machine generation or the integration with a logical planner. The state machine concept allows to quickly alter the execution by simply reconnecting some transitions. For this, no deep programming skills as required, as the GUI allows for intuitive visual programming.

Therefore, RAFCON is an ideal tool for mission control that can be used in different RoboCup competitions, for example the Logistics or Rescue League. This has been proved in the SpaceBotCup, in which similar requirements compared to the mentioned RoboCup leagues had to be met. A video of RAFCON is presented on \url{http://213.136.81.227/23483/rafcon.mp4}.

We are constantly improving RAFCON and planning to release it as open source by the end of the year via GitHub. By then, the documentation will be finished and the implementation be stable.

\section*{Acknowledgment}

This work has been funded by the Helmholtz-Gemeinschaft Germany as part of the project RACELab, by the Helmholtz Association, project alliance ROBEX, under contract number HA-304 and by the European Commission under contract number FP7-ICT-608849-EUROC.


\bibliographystyle{splncs03}
\bibliography{references}

\end{document}